\documentclass[sigconf]{acmart}

\AtBeginDocument{%
  }

\setcopyright{none}
\renewcommand\footnotetextcopyrightpermission[1]{}
\acmConference{}{}{}
\acmBooktitle{}
\acmDOI{}
\acmISBN{}

\usepackage{CJKutf8}
\newcommand{\casezh}[1]{\begin{CJK}{UTF8}{gbsn}#1\end{CJK}}

\usepackage{xcolor}
\definecolor{GoodGreen}{HTML}{00A49D}

\usepackage{array}
\usepackage{colortbl}
\colorlet{WeFairRow}{GoodGreen!6}
\colorlet{TableSection}{black!6}
\newlength{\WFAblationMethodWidth}

\newcommand{\WFmidrule}{%
  \midrule
  \noalign{\kern-\belowrulesep
    {\color{WeFairRow}\hrule height\belowrulesep}}}
\newcommand{\WFbottomrule}{%
  \noalign{{\color{WeFairRow}\hrule height\aboverulesep}%
    \kern-\aboverulesep}%
  \bottomrule}
\newsavebox{\WFHeaderReference}
\newsavebox{\WFHeaderCell}
\newcommand{\WFcenterhead}[1]{%
  \sbox{\WFHeaderReference}{\shortstack{Consistent\\Recall$\uparrow$}}%
  \sbox{\WFHeaderCell}{#1}%
  \raisebox{\dimexpr(\ht\WFHeaderReference-\dp\WFHeaderReference
    -\ht\WFHeaderCell+\dp\WFHeaderCell)/2\relax}[0pt][0pt]{%
    \usebox{\WFHeaderCell}}}
\usepackage{enumitem}
\usepackage{algorithm}     
\usepackage{algpseudocode} 

\begin{document}

\settopmatter{
  authorsperrow=3,
  printacmref=false,
  printccs=true
}

\author{Yingtao Ren}
\email{yingtaoren@tencent.com}
\authornote{These authors contributed equally to this work.}
\author{Aolin Li}
\email{churenli@tencent.com}
\authornotemark[1]
\author{Cheng Cheng}
\email{trcheng@tencent.com}
\authornotemark[1]
\affiliation{%
  \institution{Weixin Group, Tencent}
  \country{China}
}

\author{Huiyun Hu}
\email{huiyunhu@tencent.com}
\author{Huasheng Liang}
\email{watsonliang@tencent.com}
\author{Baoxing Xie}
\email{baoxingxie@tencent.com}
\affiliation{%
  \institution{Weixin Group, Tencent}
  \country{China}
}

\author{Jun Yuan}
\email{junjyuan@tencent.com}
\author{Haijun Wu}
\email{ethanhjwu@tencent.com}
\affiliation{%
  \institution{Weixin Group, Tencent}
  \country{China}
}

\author{Yiwei Fu}
\email{ywfu@sfs.ecnu.edu.cn}
\affiliation{%
  \institution{East China Normal University}
  \country{China}
}

\author{Chenxing Wang}
\email{magnuswang@tencent.com}
\authornote{Corresponding author.}
\authornotemark[1]
\affiliation{%
  \institution{Weixin Group, Tencent}
  \country{China}
}

\renewcommand{\shortauthors}{Yingtao Ren et al.}

\begin{abstract}
Short-video search needs compact query rewriters that recover users' personalized intent from ambiguous queries under strict latency constraints. Existing LLM-based methods have improved rewriting by using synthetic reasoning and subsequent consumption-behavioral feedback. However, overemphasizing subsequent consumption can favor engaging content over the user's immediate need, causing intent drift even when rewritten queries are grounded in behavioral history.
We propose WeFair, \underline{F}idelity-\underline{A}ware Al\underline{i}gnment from Human Feedback for Pe\underline{r}sonalized Query Rewriting in \underline{We}Chat short-video search. WeFair uses human feedback according to how directly it reflects personalized intent, assigning feedback fidelity to a progression from demonstration to preference to verification. We first mine actual queries and user reformulations from search sessions, and then screen them for contextual support. In the demonstration stage, a supervised fine-tuning curriculum progresses from large-scale context-to-query learning to reasoning and direct prediction on human reformulation and preservation. In the preference stage, direct preference optimization learns relative preferences from subsequent consumption feedback. In the verification stage, group relative policy optimization rewards exact agreement with reformulation or preservation targets to constrain intent drift. Further analysis shows that broad supervision benefits the full curriculum supervised training, while strict verification achieves better performance than the tested auxiliary rewards. In offline experiments, the compact WeFair-2B model outperforms strong rewriting baselines in overall accuracy and intent preservation. Furthermore, a 10-day online A/B test shows relative increases of 2.951\% in click-through rate and 2.593\% in effective video-view rate, alongside a 2.350\% relative reduction in query reformulation rate. WeFair-2B has been deployed in the WeChat short-video search system, which serves over 100 million queries per day.


\end{abstract}




\title{WeFair: Fidelity-Aware Alignment from Human Feedback for Personalized Query Rewriting in Short-Video Search
}



\maketitle

\makeatletter
\fancyhead[LE,LO]{%
  \ACM@linecountL
  \sffamily\footnotesize
  WeFair: Fidelity-Aware Alignment for Personalized Query Rewriting}
\fancyhead[RE,RO]{%
  \sffamily\footnotesize\shortauthors
  \ACM@linecountR}
\makeatother


\section{Introduction}
\begin{figure}[t] 
    \vspace{13pt}
    \centering
    \setlength{\abovecaptionskip}{0.1pt}
    \includegraphics[width=\linewidth]{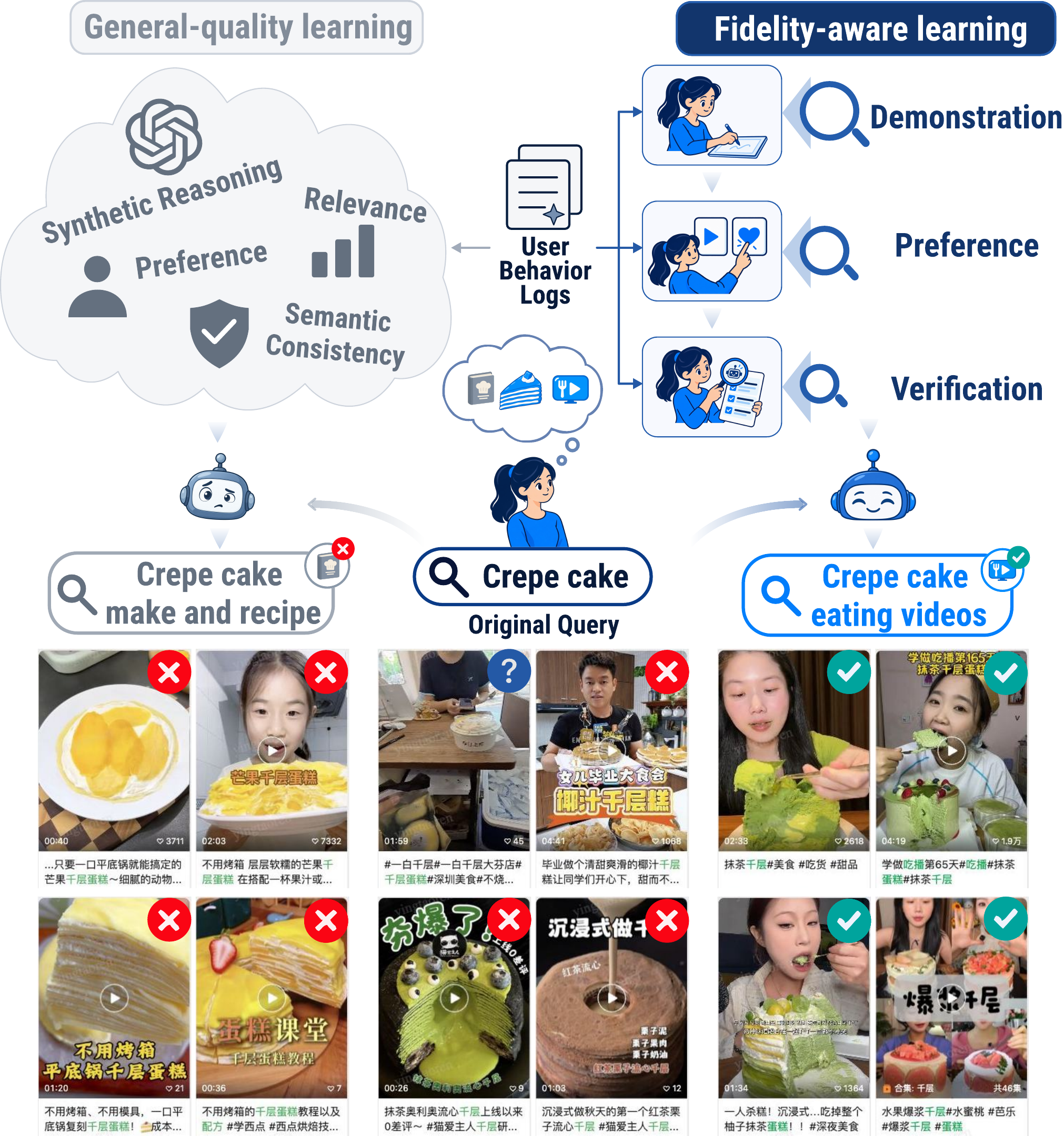}

    \caption{
        Overview of intent drift in WeChat short-video search. The original query ``crepe cake'' is redirected toward recipes instead of the intended eating videos. WeFair uses fidelity-aware human feedback through demonstration, preference, and verification to mitigate intent drift.
    }
    \label{fig:method_comparision}
    \Description{Overview of intent drift in WeChat short-video search. The original query ``crepe cake'' is redirected toward recipes instead of the intended eating videos. WeFair uses fidelity-aware human feedback through demonstration, preference, and verification to mitigate intent drift.}
\end{figure}
Personalized query rewriting is essential when an ambiguous search query does not fully express the user's immediate information need. In short-video search, the intended meaning may depend on recent searches, viewed content, or other context available when the query is issued. Rewriting can expose this information to retrieval by producing a more informative expression, but it must also satisfy strict latency constraints~\citep{gong2025cardrewriter,chu2026picque}. Therefore, the practical objective is training a compact rewrite model that can resolve ambiguity without replacing the user's current intent and needs. Furthermore, the rewrite pipeline also needs to accurately decide when to rewrite or preserve the original query. 

Although general large-size large language models (LLMs) exhibit strong reasoning capabilities, they fail to accurately understand diverse user intents and incur excessive per-inference latency, which is unacceptable for real-time search. 
To address these limitations, some approaches use synthetic Chain-of-Thought(CoT) traces to distill reasoning capabilities from large teacher LLMs into smaller rewrite models, and then use large-scale subsequent consumption feedback to align them with platform-specific user preferences.
\begin{figure*}[t]
    \centering
    
    \setlength{\abovecaptionskip}{0.1pt}
    \includegraphics[width=\textwidth]{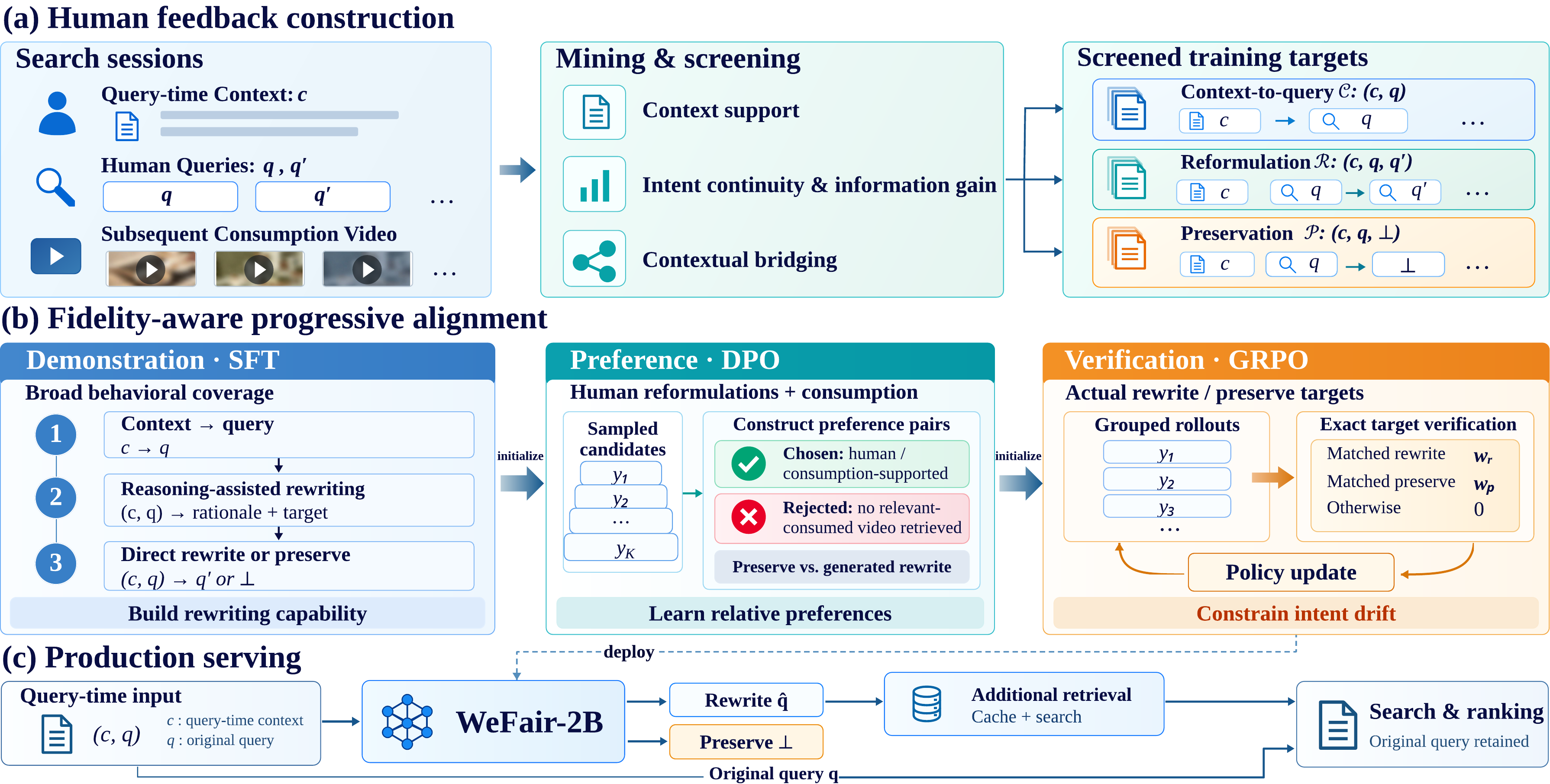}
    \vspace{-6pt}
    \caption{
WeFair training and serving. (a) Search sessions yield screened reformulation and preservation targets. (b) Progressive training aligns the model through demonstration, preference, and verification stages. (c) Rewrites expand the candidate pool through cache lookup and search while also informing online ranking.
    }
    \label{fig:overview}
    \Description{WeFair training and serving. (a) Search sessions yield screened reformulation and preservation targets. (b) Curriculum SFT builds rewriting capabilities, DPO learns preferences from reformulations and consumption feedback, and GRPO applies strict target verification to constrain intent drift. (c) Rewrites expand retrieval through cache lookup and search, while the original query remains a retrieval and ranking anchor.}
\end{figure*}

However, consumption-oriented alignment does not necessarily reflect the user's immediate information need. When insufficiently constrained by direct evidence of user intent, a rewriter may favor queries that retrieve engaging content while redirecting the original search intent, resulting in intent drift. Figure~\ref{fig:method_comparision} shows an example of intent drift. 
To better capture personalized intent, recent methods incorporate actual user reformulations, which provide more direct evidence of how users clarify their information needs. PicQue\citep{chu2026picque} uses historical reformulations as context for generating training examples that subsequently undergo manual verification, while WeWrite\citep{cheng2026wewrite} directly uses mined reformulations for supervised fine-tuning and reward construction in GRPO. While these advances demonstrate the value of real user reformulations, the subsequent consumption feedback differs in how directly they reflect personalized intent, and its fine-grained use across training stages remains underexplored.
To bridge the gap, we propose WeFair, a fidelity-aware alignment framework for training a compact personalized query rewriter model in WeChat short-video search. Its central idea is to match supervision fidelity to the requirements of each training stage, improving how human feedback guides a small model toward faithful and efficient rewriting. This progression can enable a compact model to acquire personalized query rewriting capabilities while mitigating intent drift.

WeFair combines automated data construction with three complementary training stages. We mine actual reformulations from search sessions and apply LLM screening and contextual verification to retain demonstrations grounded in user intent. Training then proceeds through three stages, each with a distinct purpose. In the demonstration stage, a supervised fine-tuning (SFT) curriculum progresses from context-to-query prediction to context-conditioned reasoning and finally direct query rewriting, building solid intent understanding capability. The broad coverage of human demonstrations prepares the compact policy model to generate useful candidates for later optimization. In the preference stage, direct preference optimization (DPO) learns from candidate pairs distinguished by subsequent consumption outcomes, increasing the probability of rewrites associated with better personalized user intent.  Finally, group relative policy optimization (GRPO) applies a strict verification reward based on exact match with actual human reformulations or preservation. This final stage anchors policy updates to observed human expressions without adding graded similarity, retrieval, or other judge rewards. At inference time, the rewrite model directly outputs a short rewrite or a preservation decision, avoiding additional CoT generation latency. Figure~\ref{fig:overview} shows the overview of the proposed WeFair.

Our further analysis suggests that the three training stages differ in their tolerance for supervision with varying intent fidelity. SFT can tolerate lower-fidelity supervision to acquire broad rewriting capabilities, while DPO can incorporate preferences inferred from subsequent consumption to refine candidate selection. In contrast, final GRPO optimization requires strict fidelity to the intended target. Auxiliary rewards may involve more fine-grained supervision but reduce rewriting accuracy. Offline and online experiments demonstrate WeFair's effectiveness in personalized rewriting. WeFair-2B rewrite model has been
deployed in the online WeChat short-video search system, serving over 100 million queries per day.
Our main contributions can be summarized as follows:
\begin{enumerate}[label=(\arabic*),labelindent=0pt,labelsep=0.5em,leftmargin=*,itemindent=0pt,align=left,topsep=2pt,partopsep=0pt,parsep=0pt,itemsep=2pt]
\item We examine the role of supervision fidelity in personalized rewriting post-training. Demonstrate that extensive broad lower-fidelity supervision benefits curriculum SFT, while strict verification better supports on-policy alignment.
\item We propose WeFair, a fidelity-aware framework that matches human feedback to demonstration, preference, and verification stages, enabling compact models to capture personalized needs while mitigating intent drift.
\item Offline and online experiments demonstrate that WeFair achieves strong rewriting performance with compact models and significant improvements in online WeChat short-video search.
\end{enumerate}

\section{Related Work}
\subsection{Personalized Query Rewriting}

LLM-based query rewriting improves retrieval by enriching query representations and adapting to search domains. Query2doc generates pseudo-documents, while CardRewriter incorporates platform knowledge through knowledge cards~\citep{wang2023query2doc,gong2025cardrewriter}. Domain adaptation further aligns query expressions with retrieval requirements. BEQUE combines supervised rewriting with retrieval feedback\citep{peng2024beque}, and negation-aware rewriting preserves explicit constraints~\citep{guo2025negation}. MAAQR and QP-OneModel extend this direction through adaptive refinement and unified query understanding~\citep{zheng2025maaqr,huang2026qponemodel}. Beyond representation enrichment, IterQR and Think-Then-Rewrite incorporate reasoning into retrieval-oriented training\citep{chen2025iterqr,li2026ttr}. MiniELM combines lightweight distillation with simulated feedback~\citep{nguyen2025minielm}. For personalized search, user context and observed behavior provide evidence of the user's current information need, helping resolve ambiguity and determine when to rewrite. PicQue combines hybrid reasoning supervision with GRPO to address over-rewriting~\citep{chu2026picque}, while Think-to-Personalize uses purchase history to support personalized retrieval~\citep{jiang2026ttp}. WeWrite mines actual user reformulations and trains with SFT and GRPO using lexical similarity and conditional query-volume rewards~\citep{cheng2026wewrite}. 

Furthermore, OneBar emphasizes content-grounded recommendations, while short-video and conversational query suggestion address consumption utility and intent coverage~\citep{tang2026onebar,tian2026clickable,liu2026intentcoverage}. DVCQR aligns separate rewrites with sparse and dense retrieval~\citep{li2026dvcqr}. These studies demonstrate the value of contextual and behavioral signals for query rewriting, while highlighting distinct objectives such as retrieval relevance, consumption utility, and personalized intent. 

\subsection{LLM Post-Training and Feedback Alignment}
Post-training methods provide different mechanisms for learning from demonstrations, preference pairs, and verified rewards. SFT learns demonstrated responses, whereas DPO optimizes chosen--rejected preferences without separately training a reward model, and GRPO estimates advantages within sampled groups~\citep{guo2025deepseek,ren_context,rafailov2023dpo,ouyang2022instructgpt}. Beyond algorithm selection, recent work examines which supervision is useful at different learning stages. Stage-specific data selection distinguishes capability acquisition from reinforcement~\citep{he2026stagespecific}. SmartSearch uses a process-reward-guided curriculum~\citep{wen2026smartsearch}. Hindsight self-distillation further derives corrections and preferences from follow-up user interactions~\citep{buening2026interactions}. This work motivates matching supervision to both the policy's capabilities and the evidence available from users.
In search applications, AWHCP combines SFT, DPO, and GRPO with quality, format, and recall rewards~\citep{ou2026awhcp}, while CoRe uses deployed relevance feedback~\citep{wen2026core}. Such signals align generation with search outcomes, but provide different evidence about intended reformulations. Comparisons under compute constraints further show that on-policy rewriting need not outperform prompted alternatives~\citep{xu2026onpolicy}, motivating examination of the optimization signals alongside the algorithms.

On-policy optimization consequently requires considering the reliability of candidate-level feedback. TTRL shows that noisy pseudo-labels can still yield reliable rewards~\citep{zuo2025ttrl}. This motivates prioritizing reward purity in on-policy preference alignment.
\section{Methodology}

\textbf{Task Formulation.}
In the short-video search system, given a query \(q\) and query-time user context \(c\), a policy \(\pi_\theta(y\mid x)\), where \(x=(c,q)\), outputs a rewritten query \(\hat q\) or a preservation marker \(\bot\). The \(\hat q\) serves as an additional retrieval query, and the preservation marker ensures that only the original query is retained. We define intent drift as \(\hat q\) changing or adding information supported by the \(c\) but differing from the user's immediate needs. Define hallucination as \(\hat q\) introducing additional information unsupported by \(x\). The objective is to generate context-supported \(\hat q\) while preserving the actual information needs. \\
\textbf{Framework Overview.}
WeFair matches supervision fidelity to the requirements of each training stage. We utilize context-to-query records, user reformulations, and subsequent consumption feedback from search sessions. These signals capture a broad range of user preferences but differ in how directly they reflect personalization and immediate intent. Accordingly, curriculum SFT builds rewriting capabilities from broad demonstrations, DPO learns relative preferences from consumption-grounded pairs, and GRPO anchors final refinement through strict agreement with screened reformulation and preservation targets.

\subsection{Human Feedback Construction} \label{sec:data_construction}
We mine and screen search-session records to construct three supervision pools: a context-to-query pool $\mathcal C=\{(c,q)\}$, a reformulation pool $\mathcal R=\{(c,q,q')\}$, and a preservation pool $\mathcal P=\{(c,q,\bot)\}$. All used data are de-identified and sanitized to protect privacy.\\
\textbf{Candidate mining.}
For context-to-query mining, we extract queries $q$ with their preceding context $c$ that have at least one effective video view. An effective video view means watching a retrieved video for more than ten seconds. We further screen all raw context-to-query records with DeepSeek-V4-Flash, retaining queries that are supported by the query-time context, yielding initial context-to-query pool $\mathcal C_0$. For reformulation mining, we link eligible adjacent queries to construct initial reformulation $\mathcal R_0$ within the same search session, requiring no video click following $q$ and an effective video view following $q'$. Each pair uses the context available when $q$ is issued.
Both records exclude suggestion-triggered queries and remove system-generated suggestions from the extracted context.
Since adjacent queries still likely express different information needs, candidate reformulations undergo rule-based filtering followed by LLM screening for contextual support and intent continuity.
Algorithm~\ref{alg:feedback_construction} summarizes the construction process. \\
\textbf{Rule-based filtering.}
After getting the initial candidates, we then conduct rule-based filtering to $\mathcal R_0$. Let $S(t)$ denote a string's character set, $a=s(q)$, $b=s(q')$, and $U=S(b)\setminus S(a)$, where $s$ removes whitespace, non-word characters, and underscores. Thus, \(U\) contains the characters introduced by the reformulation. We filter a candidate reformulation if any of the following conditions hold:
\begin{enumerate}[label=(\arabic*),labelindent=0pt,labelsep=0.3em,leftmargin=*,itemindent=0pt,align=left,topsep=2pt plus 1pt,partopsep=0pt,parsep=0pt,itemsep=0pt]  
  \item \label{rule:no-gain}\label{rule:containment}\emph{No character gain:} $U=\varnothing$.
  \item \label{rule:function-words}\emph{Only gain function words:} the added characters belong to a fixed Chinese function-word lexicon.
  \item \label{rule:overlap}\emph{Low character overlap:} the original and subsequent queries share too few characters, measured by character-set Jaccard similarity $J(a,b)=|S(a)\cap S(b)|/|S(a)\cup S(b)|\le0.06$.
  \item \label{rule:context} \emph{Insufficient contextual support:} no contextual match satisfies the criteria for new character trigrams, bigrams, single-character additions, or the common-prefix fallback.
\end{enumerate} 
These rule-based filters remove uninformative changes and reformulations with weak lexical or contextual support, reducing noise in the candidate reformulation pool and limiting the workload of subsequent LLM screening. In our practice, about half of the initial candidate reformulations are filtered out at this stage. \\
\textbf{LLM screening.}
After rule-based filtering, we further use large-size reasoning LLMs to evaluate remaining candidate reformulations. Acceptance requires intent continuity $I$, information gain $G$, and contextual bridging $B$:
\begin{equation}
K_{\mathrm{LLM}}
=\operatorname{Maj}_{j=1}^{3}[I_j(q,q')\land G_j(q,q')]
\land\operatorname{Maj}_{j=1}^{3}[B_j(q,q',c)].
\label{eq:kkk}
\end{equation}
Here, $j$ indexes the three models used at each stage, and $\operatorname{Maj}$ requires at least two accepting judgments.
Intent continuity requires the reformulation to preserve the original subject, task, answer type, and essential constraints. Information gain requires useful retrieval information or meaningful entity recovery, excluding changes that merely alter formatting. We first use three LLMs to jointly check intent continuity and information gain, $I(q,q')\land G(q,q')$, and aggregate their decisions by majority vote. Candidates passing this check proceed to a three-model majority vote on contextual bridging $B(q,q',c)$. At each stage, the models judge independently using the same input and screening criteria. This check determines whether contextual evidence supports linking the added detail to the current query. An isolated mention of the target query or a broad historical interest alone is insufficient to justify the added detail.
Each call returns acceptance or rejection decisions with a brief reason, and malformed decisions are discarded. Details are provided in Appendix~\ref{sec:imp_details}. About one third of the candidates entering LLM screening pass both checks and form the final reformulation pool \(\mathcal R\).\\
\textbf{Refuse rewrite labels.}
We construct two types of preservation examples. Natural examples come from $\mathcal C_0$ where the original query $q$ leads to at least one effective video view, providing behavioral evidence that the query retrieves useful content. Hard examples come from reformulation candidates that pass the first LLM screening stage but are rejected during contextual bridging, as the proposed additions lack support in the query-time context. These examples teach the model to avoid unsupported personalization.
We further use a reasoning LLM to verify that the query is clear and that personalization cues are ambiguous. For verified examples of both types, we retain the original context and query and assign the preservation target, yielding the final preservation pool $\mathcal P$. Finally, we remove the records selected for natural preservation from the initial context-to-query pool $\mathcal C_0$, and the remaining records form the final context-to-query pool $\mathcal C$.

\newcommand{\resdash}{\textemdash}
\newcommand{\pending}{\textit{TBD}}

\begin{table*}[!t]
\centering
\caption{Offline performance on personalized query rewriting. WeFair achieves the highest Consistent Recall and decision accuracy. WeFair-2B also exhibits lower intent drift than the larger rewrite baselines. Bold text marks the best scores.}
\label{tab:main-results}
\setlength{\tabcolsep}{\dimexpr(\textwidth-418.3191pt)/20\relax}
\begin{tabular}{@{}>{\columncolor{white}[0pt][\tabcolsep]}llcccccccc>{\columncolor{white}[\tabcolsep][0pt]}c@{}}
\toprule
& & \multicolumn{5}{c}{Rewrite Quality} & \multicolumn{3}{c}{Decision Quality} & \\
\cmidrule(lr){3-7}\cmidrule(lr){8-10}
\WFcenterhead{Method} & \WFcenterhead{Mode} & \shortstack{Consistent\\Recall$\uparrow$} & \shortstack{Consistent\\Prec.$\uparrow$} & \shortstack{EM\\Recall$\uparrow$} & \shortstack{Hallucination\\ Rate$\downarrow$} & \shortstack{Intent Drift\\Rate$\downarrow$} & \shortstack{RD\\Recall$\uparrow$} & \shortstack{PD\\Recall$\uparrow$} & \shortstack{Decision\\Acc.$\uparrow$} & \shortstack{Overall\\Acc.$\uparrow$} \\
\midrule
DeepSeek-V4.1-Flash & Full CoT & 72.92 & 64.80 & 26.06 & 1.67 & 4.28 & 80.72 & 44.42 & 67.51 & 62.55 \\
Gemini-3.1-Pro & Full CoT & 70.37 & 68.45 & \textbf{28.17} & \textbf{0.78} & \textbf{4.16} & 76.50 & 54.02 & 68.32 & 64.42 \\
Qwen3.8-27B & Full CoT & 71.03 & 62.14 & 26.53 & 1.77 & 5.09 & 79.82 & 39.73 & 65.23 & 59.64 \\
Qwen3.8-27B & Direct & 50.32 & 61.12 & 17.72 & 1.92 & 6.13 & 61.13 & \textbf{62.95} & 61.79 & 54.91 \\
\midrule
WeWrite-0.8B & Direct & 66.79 & 51.88 & 17.84 & 5.47 & 13.34 & 89.27 & 31.03 & 68.07 & 53.77 \\
WeWrite-2B & Direct & 68.58 & 46.57 & 18.54 & 6.39 & 16.45 & 95.79 & 10.04 & 64.59 & 47.28 \\
WeWrite-4B & Direct & 70.37 & 50.27 & 21.83 & 3.83 & 15.88 & 94.38 & 20.31 & 67.42 & 52.15 \\
AWHCP-2B & Fast CoT  & 65.77 & 43.28 & 19.72 & 7.28 & 14.88 & \textbf{97.32} & 4.46 & 63.53 & 43.46 \\
AWHCP-8B & Fast CoT  & 71.58 & 49.65 & 21.71 & 6.70 & 13.34 & 94.64 & 13.39 & 65.07 & 50.41 \\
\midrule
WeFair-CoT-2B & Fast CoT & 81.12 & 69.15 & 26.77 & 2.33 & 7.25 & 92.66 & 56.89 & 79.64 & 72.30 \\
WeFair-CoT-4B & Fast CoT & 81.25 & 68.76 & 27.14 & 2.64 & 6.91 & 93.28 & 56.52 & 79.90 & 72.25 \\
\WFmidrule
\rowcolor{WeFairRow}
WeFair-0.8B & Direct & 76.88 & 64.01 & 26.53 & 1.79 & 9.45 & 91.19 & 49.46 & 76.00 & 66.90 \\
\rowcolor{WeFairRow}
WeFair-2B & Direct & \textbf{83.95} & 69.26 & 26.29 & 2.43 & 5.36 & 94.51 & 53.35 & 79.53 & 72.81 \\
\rowcolor{WeFairRow}
WeFair-4B & Direct & 83.93 & \textbf{70.08} & 26.17 & 2.17 & 6.03 & 94.13 & 55.20 & \textbf{79.97} & \textbf{73.48} \\
\WFbottomrule
\end{tabular}
\vspace{2pt}
\end{table*}

\subsection{Fidelity-Aware Progressive Alignment}\label{sec:method_alignment}
\begin{algorithm}[t]
\caption{Human feedback construction}
\label{alg:feedback_construction}
\small
\begin{algorithmic}[1]
\Require Search-session records $\mathcal S$
\Ensure Context-to-query $\mathcal C$, reformulation $\mathcal R$,
        preservation $\mathbf P$

\State \mbox{Mine context--query records $\mathcal C_0$
       and adjacent-query pairs $\mathcal R_0$}
\State \mbox{Exclude suggestion-triggered queries; clean contexts}
\State \mbox{Screen $\mathcal C_0$ for contextual support;
       filter $\mathcal R_0$ by rules}
\State \mbox{$\mathcal N\gets$ natural preservation records
       selected from $\mathcal C_0$}
\State \mbox{$\mathcal C\gets\mathcal C_0\setminus\mathcal N$;
       $\mathcal R\gets\varnothing$;
       $\mathbf P\gets\{(c,q,\bot)\mid(c,q)\in\mathcal N\}$}
\For{each $(c,q,q')\in\mathcal R_0$}
    \State \mbox{$d_1\gets$ majority vote of three $I\land G$ judges}
    \State \mbox{\textbf{if} $d_1=\mathrm{reject}$ \textbf{then continue}}
    \State \mbox{$d\gets$ majority vote of three context judges}
    \State \mbox{\textbf{if} $d=\mathrm{accept}$ \textbf{then}
           $\mathcal R\gets\mathcal R\cup\{(c,q,q')\}$}
    \State \mbox{\textbf{if} $d=\mathrm{reject}$ \textbf{then}
           $\mathbf P\gets\mathbf P\cup\{(c,q,\bot)\}$}
\EndFor
\State $\mathbf P\gets\Call{VerifyPreservation}{\mathbf P}$
\State \Return $\mathcal C,\mathcal R,\mathbf P$
\end{algorithmic}
\end{algorithm}
The training of WeFair follows demonstration, preference, and verification, with each stage initializing the next. \\
\textbf{Demonstration: curriculum SFT.}
The curriculum consists of three phases. Phase~1 trains the model to predict the query $q$ from preceding context $c$. Phase~2 takes $x=(c,q)$ as input and learns to generate a
teacher-provided rationale followed by the final target $y^*$.
Phase~3 predicts $y^*$ directly from $x$. Phase~1 uses context-to-query pool $\mathcal{C}$, while Phases~2 and~3 use a small part from reformulation pool $\mathcal{R}$ and preservation pool $\mathcal{P}$. $\mathcal{C}$ provides broad behavioral coverage, while the smaller adaptation set teaches evidence-supported rewriting, preservation, and the serving format. The rationales are auxiliary explanations generated by Gemini-3.1-Pro-Preview. For each phase $k\in\{1,2,3\}$, let
$\mathcal D_{\mathrm S}^{(k)}$ denote the corresponding
training set of input--target pairs $(u,v)$.
We optimize the standard objective:
\begin{equation}
\mathcal L_{\mathrm{SFT}}^{(k)}(\theta)
=
-\mathbb E_{(u,v)\sim\mathcal D_{\mathrm S}^{(k)}}
\left[\log \pi_\theta(v\mid u)\right].
\label{eq:wefair-sft}
\end{equation}
Each phase continues from the preceding checkpoint. We obtain $\pi_{\mathrm{SFT}}$ after the final direct-prediction phase.\\
\textbf{Preference: DPO with consumption pairs.}
Starting from $\pi_{\mathrm{SFT}}$, we generate 64 candidate
responses for each input $x$ and construct up to two DPO pairs. Using up to three subsequent relevant consumed videos as user preference, a chosen candidate \(y^+\) must have at least one relevant consumed video appear in its top-10 retrieval results. A rejected candidate \(y^-\) satisfies no relevant consumed video in its top-100 retrieval results. The first pair uses the human reformulation \(q'\) as the positive, yielding \((x,q',y^-_1)\). The second pair uses a consumption-supported chosen candidate, yielding \((x,y^+_{\mathrm{cons}},y^-_2)\). If only one valid negative candidate is available, we set \(y^-_1=y^-_2\) and share it across both pairs. For preservation-required examples, we pair the preservation marker $\bot$ with a generated rewrite $y^-$, yielding $(x,\bot,y^-)$. DPO increases the relative likelihood of human reformulations or consumption-supported candidates over observed low-engagement alternatives. Because engagement can differ from the immediate information need, the final verification stage returns to explicit reformulation and preservation markers.\\
\textbf{Verification: human-grounded GRPO.}
Verification is the final mechanism for constraining optimization to direct target evidence with GRPO. Starting from the DPO policy, we sample 32 responses for each $(c,q)$. After deterministic parsing, a valid response $y$ receives reward:
\begin{equation}
R(y,y^*)=
\begin{cases}
w_{\mathrm r}, & y=y^*,\quad y^*\ne\bot,\\
w_{\mathrm p}, & y=y^*=\bot,\\
0, & \text{otherwise},
\end{cases}
\label{eq:wefair-reward}
\end{equation}
where $w_{\mathrm r},w_{\mathrm p}>0$ weight exact matching with the retained human query $q'$ or preservation marker $\bot$, and any other responses receive zero reward, ensuring the highest fidelity signal supervision. Our further analysis shows that the high-fidelity reward function outperforms other designs with graded similarity, retrieval, or learned-judge rewards.
This conservative final stage anchors updates to human reformulations and preservation labels, mitigating consumption-driven intent drift. Notably, generating a preservation marker is easier than generating
an accurate rewritten query, so the policy may engage in reward hacking by
excessively favoring preservation. We therefore design a concise
data ratio and reward assignment to mitigate this risk, with implementation details provided in Appendix~\ref{sec:imp_details}.


\subsection{Integration into Retrieval and Ranking}
\label{sec:online-deployment}
We integrate rewritten queries $\hat q$ into retrieval and ranking through two complementary paths. A lightweight fake-recall path obtains cached candidates associated with $\hat q$, following WeWrite~\citep{cheng2026wewrite}, while an additional lightweight vector retrieval uses $\hat q$ to further expand the candidate pool. The rank system uses video relevance to both the original query $q$ and $\hat q$. We gate the computation of rewrite-based relevance using the probability of the preservation marker token $\bot$ at the first decoding step. When this probability exceeds the threshold, we compute relevance using only $q$. Otherwise, we compute relevance scores with respect to both $q$ and $\hat q$ and use their weighted combination in the final video ranking. This design retains the original query as a ranking anchor while allowing contextual refinement. Appendix~\ref{sec:imp_details} provides further details.

\section{Experiment}

\subsection{Experimental Setup}
\label{sec:experimental-setup}

\textbf{Offline evaluation.}
We trained our method on four weeks of search logs and evaluated all methods on the test set from the following three days. Let $N_R$ and $N_P$ denote the numbers of rewrite-required and preservation-required examples in the test set. During the evaluation, all the evaluated rewriters use the original input query and query-time context. We use DeepSeek-V4.1-Flash as the LLM judge, which additionally considers preceding and subsequent user behavior to assess whether a generated query is consistent with the information need expressed by the user's actual subsequent reformulation. To test the effectiveness of the judge, we randomly sampled 500 responses of Gemini-3.1-Pro rewriter from the test set and evaluated each response using both the LLM judge and human review\citep{chu2026picque}. The LLM judge showed 96.6\% agreement with human judgments, with Cohen’s Kappa = 0.919.\\
\textbf{Baselines and implementation.} For general LLM baselines, we evaluate DeepSeek-V4.1-Flash, Gemini-3.1-Pro, and Qwen3.8-27B. Furthermore, we reproduce two other adaptive rewrite methods, WeWrite and AWHCP, using the same training set. We reproduced WeWrite following the complete procedure described in its original paper~\citep{cheng2026wewrite}, and adapted AWHCP~\citep{ou2026awhcp} to the WeChat search system (Details in Appendix~\ref{sec:imp_details}). For our method, WeFair was evaluated at Qwen3.5-0.8B, Qwen3.5-2B, and Qwen3-4B. WeFair-CoT introduces thinking at the GRPO training stage with up to 64 output tokens, and only assesses the resulting rewrite quality. \\
\textbf{Evaluation Metrics.}
We mainly evaluate rewrite and decision-making quality of all baselines. Specifically, let $C$ be the count of judge-accepted rewrites on the rewrite-required set, $W$ be all triggered rewrites, $W_R$ be the correct triggered rewrites decision, and $P_C$ be the correct preservations. \emph{Consistent Recall} and \emph{Precision} metrics are $C/N_R$ and $C/W$, respectively. \emph{Exact match (EM)} counts rewrites identical to the actual human subsequent reformulation, divided by $N_R$. On rewrite-required examples, \emph{Hallucination Rate} counts triggered rewrites containing context-unsupported additions, whereas \emph{Intent Drift Rate} counts context-supported rewrites that fail to match the user's current information need. Both rates are divided by $N_R$. \emph{Rewrite-decision (RD) recall} is $W_R/N_R$, and \emph{preservation-decision (PD) recall} is $P_C/N_P$. \emph{Decision accuracy} is $(W_R+P_C)/N$, whereas \emph{overall accuracy} is $(C+P_C)/N$. All scores are percentages. More details about setting and implementation are shown in Appendix~\ref{sec:imp_details}

\begin{table}[t]
\centering
\caption{Ablation of our curriculum SFT at 2B. The full curriculum SFT achieves the highest consistency recall and overall accuracy. Bold text marks the best scores.}
\label{tab:sft-curriculum}
\settowidth{\WFAblationMethodWidth}{Full Curriculum}
\setlength{\tabcolsep}{\dimexpr(\columnwidth-226.4666pt)/12\relax}
\begin{tabular}{@{}>{\columncolor{white}[0pt][\tabcolsep]}lrrrrr>{\columncolor{white}[\tabcolsep][0pt]}r@{}}
\toprule
\WFcenterhead{\makebox[\WFAblationMethodWidth][l]{Training}} & \multicolumn{1}{c}{\shortstack{Consistent\\Rec.$\uparrow$}} & \multicolumn{1}{c}{\shortstack{Consistent\\Prec.$\uparrow$}} & \multicolumn{1}{c}{\shortstack{RD\\Rec.$\uparrow$}} & \multicolumn{1}{c}{\shortstack{PD\\Rec.$\uparrow$}} & \multicolumn{1}{c}{\shortstack{Dec.\\Acc.$\uparrow$}} & \multicolumn{1}{c@{}}{\shortstack{Overall\\Acc.$\uparrow$}} \\
\midrule
C2Q$\to$HR & 69.45 & 46.40 & \textbf{97.28} & 8.38 & 64.93 & 47.23 \\
CoT$\to$HR & 66.76 & 48.62 & 93.93 & 24.18 & 68.55 & 51.26 \\
HR Only & 66.70 & 47.29 & 91.26 & 13.02 & 62.78 & 47.16 \\
\rowcolor{WeFairRow}
Full Curriculum & \textbf{72.67} & \textbf{53.08} & 94.25 & \textbf{25.45} & \textbf{69.21} & \textbf{55.48} \\
\WFbottomrule
\end{tabular}
\vspace{4pt}
\centering
\caption{Ablation of progressive alignment stages at 2B with equal total training data. The full WeFair sequence achieves the highest consistent recall and overall accuracy. Bold text marks the best scores.}
\label{tab:stage-ablation}
\setlength{\tabcolsep}{\dimexpr(\columnwidth-226.4666pt)/12\relax}
\begin{tabular}{@{}>{\columncolor{white}[0pt][\tabcolsep]}lrrrrr>{\columncolor{white}[\tabcolsep][0pt]}r@{}}
\toprule
\WFcenterhead{\makebox[\WFAblationMethodWidth][l]{Training}} & \multicolumn{1}{c}{\shortstack{Consistent\\Rec.$\uparrow$}} & \multicolumn{1}{c}{\shortstack{Consistent\\Prec.$\uparrow$}} & \multicolumn{1}{c}{\shortstack{RD\\Rec.$\uparrow$}} & \multicolumn{1}{c}{\shortstack{PD\\Rec.$\uparrow$}} & \multicolumn{1}{c}{\shortstack{Dec.\\Acc.$\uparrow$}} & \multicolumn{1}{c@{}}{\shortstack{Overall\\Acc.$\uparrow$}} \\
\midrule
SFT Only & 72.67 & 53.08 & 94.25 & 25.45 & 69.21 & 55.48 \\
SFT$\to$DPO & 75.86 & 54.25 & \textbf{95.65} & 22.78 & 69.13 & 56.54 \\
SFT$\to$GRPO & 78.93 & 66.43 & 93.28 & \textbf{55.37} & 79.48 & 70.35 \\
\rowcolor{WeFairRow}
WeFair & \textbf{83.95} & \textbf{69.26} & 94.51 & 53.35 & \textbf{79.53} & \textbf{72.81} \\
\WFbottomrule
\end{tabular}
\end{table}

\begin{table*}[t]
\centering
\caption{Rewriting performance and advantage purity under different GRPO rewards. All variants start from SFT without DPO and use equal total training data. Bold text marks the best scores.}
\label{tab:reward-analysis}
\setlength{\tabcolsep}{\dimexpr(\textwidth-384.08319pt)/18\relax}
\begin{tabular}{@{}>{\columncolor{white}[0pt][\tabcolsep]}lcccccccc>{\columncolor{white}[\tabcolsep][0pt]}c@{}}
\toprule
& \multicolumn{2}{c}{Rewrite quality} & \multicolumn{3}{c}{Advantage purity} & \multicolumn{3}{c}{Decision quality} & \\
\cmidrule(lr){2-3}\cmidrule(lr){4-6}\cmidrule(lr){7-9}
\WFcenterhead{Reward variant} & \shortstack{Consistent\\Recall$\uparrow$} & \shortstack{EM\\Recall$\uparrow$} & \WFcenterhead{$P_+\uparrow$} & \WFcenterhead{$P_-\uparrow$} & \WFcenterhead{$P_{\mathrm{all}}\uparrow$} & \shortstack{RD\\Recall$\uparrow$} & \shortstack{PD\\Recall$\uparrow$} & \shortstack{Decision\\Acc.$\uparrow$} & \shortstack{Overall\\Acc.$\uparrow$} \\
\midrule
Similarity-Augmented Verification & 65.39 & 18.11 & 82.98 & 61.64 & 72.13 & 94.89 & 13.81 & 65.38 & 46.62 \\
Consumption-Augmented Verification & 63.47 & 18.52 & 70.29 & 65.57 & 67.60 & 91.06 & 15.85 & 63.69 & 46.14 \\
RM-Augmented Verification & 68.20 & 16.86 & 72.30 & \textbf{76.32} & 74.05 & \textbf{98.47} & 14.24 & 67.82 & 48.56 \\
RM-Augmented Verification(4 Vote) & 69.99 & 19.54 & 64.46 & 76.27 & 70.55 & 95.53 & 17.37 & 67.08 & 50.84 \\
\WFmidrule
\rowcolor{WeFairRow}
Strict Verification & \textbf{78.93} & \textbf{23.00} & \textbf{100.00} & 71.50 & \textbf{83.75} & 93.28 & \textbf{55.37} & \textbf{79.48} & \textbf{70.35} \\
\WFbottomrule
\end{tabular}
\end{table*}

\subsection{Overall Offline Performance}
\label{sec:overall-performance}
We present the offline performances in Table~\ref{tab:main-results}. WeFair achieves the best overall performance. The WeFair-4B model reaches 73.48\% overall accuracy, exceeding the strongest large-size baselines Gemini3.1 and DeepSeek4.1. WeFair-2B achieves 72.81\% and the highest consistent Recall of 83.95\%. Its EM also exceeded the other two adaptive rewrite baselines. Gemini~3.1 retained the highest EM and the lowest Hallucination and Intent Drift Rates, while achieving much lower rewrite recall. This suggests that large-size models with CoT can substantially reduce hallucination and intent drift, but yield more incorrect preservation. WeFair's advantage lies in the balance of successful rewriting and preservation. Notably, our small-size WeFair-2B achieves significantly lower hallucination and intent drift rates than other comparable rewriting methods (with its intent drift rate reduced to about one-third), validating the effectiveness of our fidelity-aware training approach. \\
\textbf{Intent fidelity.}
WeFair achieves lower hallucination and intent drift rates than WeWrite and AWHCP, while more reliably preserving queries that did not require rewriting. Both baselines tended to favor rewriting, particularly AWHCP. Even the smallest WeFair-0.8B surpasses AWHCP-8B in preservation. The large-size LLMs generally had lower intent drift rates than WeFair but also lower successful-rewrite recall. These trends support the value of fidelity-aware alignment with human feedback in mitigating intent drift while maintaining effective personalized rewriting.\\

\textbf{Gains from model scaling.}
WeFair's overall accuracy increased by 5.91 percentage points from 0.8B to 2B, but only 0.67 points from 2B to 4B, while consistent recall remained nearly unchanged. These results indicate diminishing gains over the backbone sizes, with most of the observed gain already attained at 2B. \\
\textbf{Inference mode.}
Direct WeFair exceeds WeFair-CoT in overall accuracy at both 2B and 4B, and also achieves lower intent drift. CoT variants yield higher EM and PD Recall, but these gains did not translate into better overall performance, indicating explicit reasoning is unnecessary for the strongest results in these comparisons. For general large-size LLMs, we observe that Qwen3.8-27B with CoT achieves much better performance than direct prediction. These results suggest that our fidelity-aware training enables the rewriter to perform more effective implicit reasoning than explicit CoT. We further analyze the corresponding decode tokens and inference latency in Section~\ref{sec:efficiency}.

\begingroup
\makeatletter
\renewcommand\subsection{\def\@toclevel{2}%
  \@startsection{subsection}{2}{\z@}%
  {-6pt \@plus -.5pt \@minus -.2pt}%
  {.25\baselineskip}%
  {\ACM@NRadjust\@subsecfont}}
\renewcommand\subsubsection{\def\@toclevel{3}%
  \@startsection{subsubsection}{3}{\z@}%
  {-4pt \@plus -.5pt \@minus -.2pt}%
  {-3.5\p@}%
  {\ACM@NRadjust{\@subsubsecfont\@adddotafter}}}
\makeatother
\setlength{\parskip}{0pt plus 1pt}
\subsection{Ablation Study}
\label{sec:stage-analysis}
\subsubsection{Ablation Study of Curriculum SFT} 
To examine the role of curriculum learning in the SFT stage, we ablate the C2Q and CoT phases before direct human-target adaptation (HR). All four variants use the same number of human reformulation examples, but C2Q adds context-to-query samples in $C$. The results in Table~\ref{tab:sft-curriculum} show that this broader, lower-fidelity demonstration supervision is most beneficial when combined with the subsequent human-target learning phases. Specifically, C2Q$\to$HR improves consistent recall over HR Only but reduces preservation recall, producing little overall gain. CoT$\to$HR achieves higher preservation recall and decision accuracy than both variants, supporting the contribution of reasoning supervision to rewrite-or-preserve decisions. Full Curriculum further improves rewrite consistency and decision accuracy, achieving the highest overall accuracy. These findings suggest that the broad and lower-fidelity C2Q and CoT supervision can aid capability acquisition in our curriculum SFT.

\vspace{2pt}
\subsubsection{Ablation Study of Progressive Alignment.}
To assess the contribution of each stage in our progressive alignment framework, we compared four training configurations on the 2B model: SFT Only, SFT$\to$DPO, SFT$\to$GRPO, and the full training sequence (WeFair). All configurations use the same total amount of training data. Results are shown in Table~\ref{tab:stage-ablation}.

\noindent\textbf{Demonstration builds the rewriting foundation.}
Our curriculum SFT already surpasses all evaluated adaptive baseline variants in consistent recall. This supports the effectiveness of demonstration-based learning in establishing rewriting capabilities in compact models. However, its low PD recall reveals a strong tendency to rewrite queries even when preservation is appropriate. Demonstration learning provides a capable starting point but remains insufficient for reliable rewrite-or-preserve decisions.

\noindent\textbf{Consumption preference alignment further improves rewriting.}
The subsequent consumption-guided DPO yields modest improvements in consistent recall and precision, further strengthening successful rewriting. However, PD Recall decreases slightly, also indicating an increased tendency to rewrite preservation-required queries. This imbalance limits the improvement in overall accuracy, suggesting that consumption-based preference alignment refines rewriting capabilities but cannot reduce over-writing.

\noindent\textbf{Verification delivers the largest benefits.}
Adding strict-verification GRPO produces the largest improvement in overall accuracy, substantially improving preservation while maintaining high rewrite-decision recall. Importantly, the full sequence achieves higher consistent recall and precision than SFT$\rightarrow$GRPO. These gaps even exceed the corresponding gains from adding DPO to SFT alone, suggesting that the value of preference alignment extends beyond its immediate improvements. Although SFT$\rightarrow$GRPO achieves slightly higher PD recall, the full sequence delivers the best overall performance by combining successful rewriting with effective preservation.
All these ablation results support the complementary roles of demonstration, preference, and verification. Their progressive combination achieves the strongest overall performance.

\par\endgroup
\begingroup
\makeatletter
\renewcommand\subsection{\def\@toclevel{2}%
  \@startsection{subsection}{2}{\z@}%
  {-2pt}
  {1pt}
  {\ACM@NRadjust\@subsecfont}}
\makeatother
\setlength{\parskip}{0pt plus .1pt}
\setlength{\textfloatsep}{2pt}

\begin{table}[t]
\centering
\caption{Inference efficiency across baselines and our methods. Token counts and latencies are per-sample means, including reasoning tokens. Bold text marks the best scores.}
\label{tab:efficiency}
\renewcommand{\arraystretch}{0.96}
\setlength{\tabcolsep}{\dimexpr(\columnwidth-224.65752pt)/6\relax}
\begin{tabular}{@{}>{\columncolor{white}[0pt][\tabcolsep]}llr>{\columncolor{white}[\tabcolsep][0pt]}r@{}}
\toprule
Method & Mode & \multicolumn{1}{c}{\shortstack{Decoded tokens$\downarrow$}} & \multicolumn{1}{c@{}}{\shortstack{Mean time(s)$\downarrow$}} \\
\midrule
DeepSeek-V4.1-Flash & CoT & 1340.36 & \multicolumn{1}{c@{}}{\resdash} \\
Gemini-3.1-Pro & CoT & 474.97 & \multicolumn{1}{c@{}}{\resdash} \\
Qwen3.8-27B & CoT & 1023.38 & 61.380 \\
Qwen3.8-27B & Direct & 8.49 & 1.094 \\
\midrule
WeWrite-0.8B & Direct & 8.46 & 0.327 \\
WeWrite-2B & Direct & 8.77 & 0.331 \\
WeWrite-4B & Direct & 8.92 & 0.426 \\
AWHCP-2B & Fast CoT & 45.60 & 1.172 \\
AWHCP-8B & Fast CoT & 47.06 & 1.687 \\
\midrule
WeFair-CoT-2B & Fast CoT & 26.65 & 0.795 \\
WeFair-CoT-4B & Fast CoT & 27.98 & 0.965 \\
\WFmidrule
\rowcolor{WeFairRow}
WeFair-0.8B & Direct & \textbf{7.48} & \textbf{0.314} \\
\rowcolor{WeFairRow}
WeFair-2B & Direct & 7.60 & 0.316 \\
\rowcolor{WeFairRow}
WeFair-4B & Direct & 7.65 & 0.416 \\
\WFbottomrule
\end{tabular}
\end{table}

\begin{table*}[!t]
\centering
\caption{Selected online cases of personalized rewriting and retrieval. Control denotes the comparison system for each case. Better and Degrade markers indicate improved and degraded alignment with the inferred intent.}
\label{tab:gsb-cases}
\small
\setlength{\tabcolsep}{4pt}
\renewcommand{\arraystretch}{1.08}
\begin{tabular}{@{}
>{\raggedright\arraybackslash}m{\dimexpr0.13\textwidth-1.04\tabcolsep\relax}
>{\raggedright\arraybackslash}m{\dimexpr0.16\textwidth-1.28\tabcolsep\relax}
>{\raggedright\arraybackslash}m{\dimexpr0.25\textwidth-2\tabcolsep\relax}
>{\raggedright\arraybackslash}m{\dimexpr0.23\textwidth-1.84\tabcolsep\relax}
>{\raggedright\arraybackslash}m{\dimexpr0.23\textwidth-1.84\tabcolsep\relax}@{}}
\toprule
\textbf{Original\newline query} & \textbf{Rewritten\newline query} & \textbf{Query-time context\newline and inferred intent} & \textbf{Control retrieval strategy\newline returned results} & \textbf{WeFair retrieval strategy\newline returned results} \\
\midrule
\texttt{anggwe} & \texttt{angew} & Consumption history included ``Angew. Chem. Int. Ed.'' and materials-science articles. Inferred need: Angew-related research. & Engwe e-bike brand videos at VP1--3. & X-MOL research sharing at VP1 and Angew research videos at VP2; an Engwe result remains at VP3. \textcolor{GoodGreen}{\textbf{<Better>}}\\
\casezh{推动高质量发展}\newline(Promoting high-quality development) & \casezh{推动高质量发展 意义}\newline(Significance of promoting high-quality development) & Prior search: ``how to answer significance questions in civics''. Inferred need: an educational explanation. & Reform news at VP1,  industry and regional development videos at VP2--3. & Significance explainers at VP1--2, and civics textbook analysis at VP3. \textcolor{GoodGreen}{\textbf{<Better>}}\\
\casezh{杭州外国语学校}\newline(Hangzhou Foreign Languages School) & \casezh{杭州外国语学校招聘}\newline(Hangzhou Foreign Languages School recruitment) & Recent watch included ``Meitu recruitment'' and other employers' hiring notices. Inferred need: school recruitment. & School video at VP1, an encyclopedia video at VP2, and exam results and enrollment advice at VP3. & School video at VP1, teacher recruitment notice at VP2, and proposed-appointment notice at VP3. \textcolor{GoodGreen}{\textbf{<Better>}}\\
\casezh{教育号}\newline(Education account) & \casezh{教育局视频号}\newline(Education Bureau video account) & Prior searches: XX City Education Bureau account; a recorded click reached XX City Education. Inferred need: revisit that account. & Videos from the target account at VP1--2, unrelated education content at VP3. & Provincial/Guangzhou education content at VP1--3. Target XX City account absent from VP1--6. \textcolor{red}{\textbf{<Degrade>}}\\
\bottomrule
\end{tabular}
\vspace{1pt}
\end{table*}

\begin{table}[t]
\centering
\caption{Online A/B test results over 10 days. CTR, VV>10s, and QRR are relative changes versus the control and are statistically significant (P<0.01).} 
\label{tab:online-results}
\begin{tabular*}{\columnwidth}{@{\extracolsep{\fill}}lccc@{}}
\toprule
Strategy & CTR & VV>10s & \shortstack{QRR}\\
\midrule
Cache Retrieval & +1.954\% & +2.281\% & -1.628\%\\
Cache + Vector Retrieval & +2.951\% & +2.593\% & -2.350\%\\

\bottomrule
\end{tabular*}
\end{table}

\subsection{On-policy Reward Purity Analysis}
\label{sec:reward-analysis}
To examine whether auxiliary rewards improve on-policy alignment beyond the strict target verification, we compare five GRPO reward configurations initialized from the same startpoint. $\pi_{\mathrm{SFT}}$. All configurations use the same total amount of training data. Strict Verification(SV) uses only the reward in Eq.~\eqref{eq:wefair-reward}, rewarding exact agreement with a screened human reformulation or the preservation target. The augmented variants retain this reward while adding semantic similarity (Similarity-Augmented Verification), posterior consumption feedback (Consumption-Augmented Verification), or reward-model feedback (RM-Augmented Verification). The consumption signal follows the criteria used for DPO preference construction. We further evaluate vote-aggregated RM feedback (RM-Augmented Verification 4 Vote), where four independent decisions are obtained from four rollouts of the same RM, and a response is accepted only if all four decisions accept it. Given the computational and data construction cost of LLM-based judging, we estimate advantage purity on 500 sampled queries using the same judging procedure as in the offline benchmark. Let $z_{ij}\in\{0,1\}$ indicate the judged correctness of candidate $j$ in group $i$, and let $A_{ij}$ denote its GRPO advantage. Define $\mathcal S_+=\{(i,j):A_{ij}>0\}$ and $\mathcal S_-=\{(i,j):A_{ij}<0\}$. Positive purity measures the fraction of favored candidates judged correct, while negative purity measures the fraction of discouraged candidates judged incorrect:

\begingroup
\setlength{\abovedisplayskip}{4pt plus .1pt}
\setlength{\belowdisplayskip}{4pt plus .1pt}
\setlength{\abovedisplayshortskip}{2pt plus .1pt}
\setlength{\belowdisplayshortskip}{4pt plus .1pt}
\begin{equation}
\begin{aligned}
&\resizebox{0.98\columnwidth}{!}{$\displaystyle
P_+=\frac{\sum_{(i,j)\in\mathcal S_+}z_{ij}}
{|\mathcal S_+|},
\enspace
P_-=\frac{\sum_{(i,j)\in\mathcal S_-}(1-z_{ij})}
{|\mathcal S_-|},
\enspace
P_{\mathrm{all}}=
\frac{|\mathcal S_+|P_+ + |\mathcal S_-|P_-}
{|\mathcal S_+|+|\mathcal S_-|}.
$}
\end{aligned}
\label{eq:advantage-purity}
\end{equation}
\endgroup
Overall purity aggregates both sets, excluding zero-advantage candidates. All the results are reported in Table~\ref{tab:reward-analysis}.\\
\textbf{Strict verification provides more reliable positive reinforcement.}
SV reward pattern achieves the highest positive and overall advantage purity, together with the strongest overall accuracy among the evaluated configurations. Adding RM feedback reduces positive purity and substantially weakens preservation, suggesting that auxiliary reward models provide less reliable positive reinforcement than strict target agreement. SV has lower negative purity than RM-Augmented Verification, consistent with exact matching discouraging some acceptable alternative reformulations. However, according to the SV setting, whenever certain responses are assigned negative advantage, there always is a perfectly matching response that provides a positive reward. Consequently, the impact of incorrect negative advantage on the SV model is likely to be weakened. These results support prioritizing high-fidelity verification in final on-policy alignment without implying perfect semantic supervision. The RM-Augmented Verification 4 Vote variant improves overall accuracy over RM-Augmented Verification but lowers both positive and overall advantage purity. Four Vote setting over repeated judgments from the same RM therefore does not automatically make the induced advantage signals more reliable. This discrepancy also shows that advantage purity is a diagnostic of training-signal reliability.


\subsection{Offline Efficiency Analysis}
\label{sec:efficiency}
To examine the effects of model size and inference mode, we measure the mean number of decoded tokens and per-sample inference latency with the same computing resource. The DeepSeek and Gemini baselines use the API call and only report the mean number of decoded tokens. The other methods are deployed on an Nvidia H20 GPU with the same settings. Decoded tokens include reasoning when CoT is enabled. All the results are reported in Table~\ref {tab:efficiency}. The details of the implementation are described in Appendix~\ref{sec:imp_details}. 

General-purpose LLMs achieve competitive rewriting performance without task-specific fine-tuning, but their full-CoT configurations incur substantial decoding overhead. The CoT--Direct comparison for Qwen3.8-27B illustrates this trade-off. Explicit reasoning improves rewriting quality but sharply increases inference time, making full-CoT inference impractical in real-time search systems. 
In contrast, WeFair achieves low latency while retaining strong rewrite quality. WeFair-2B required only 7.60 decoded tokens and 0.316~s per sample, compared with 26.65 tokens and 0.795~s for WeFair-CoT-2B. At 4B size, direct WeFair used 72.66\% fewer tokens and achieved a $2.32\times$ speedup over its CoT variant. Notably, both direct variants also attained higher offline overall accuracy than CoT variants(See Table~\ref{tab:main-results}). Relative to WeWrite at matched sizes, WeFair delivered higher overall accuracy and slightly lower mean latency.
For the three direct WeFair variants, WeFair-2B retained nearly the latency of WeFair-0.8B while improving overall accuracy. Increasing the size to 4B significantly raises inference time with only a slight accuracy gain. Accordingly, we select WeFair-2B for both online A/B testing and production deployment.

\par\endgroup
\vspace{-5pt}

\subsection{Online Evaluation} \label{sec:online_eva}
To evaluate WeFair's online performance, we conducted a 10-day online A/B test in WeChat short-video search, deploying WeFair-2B on 16 NVIDIA L20 GPUs. The production control method is WeWrite\citep{cheng2026wewrite}. In production, WeFair generates rewrites for 42.26\% of processed queries and outputs preservation markers for 57.74\%. We measured click-through rate (CTR), effective video-view rate (VV>10s), and query reformulation rate (QRR). We also measured P95 model inference latency (81 ms) and the end-to-end search latency increase (+16 ms).
As shown in Table~\ref{tab:online-results}, WeFair with cache + vector retrieval strategy increases CTR and VV>10s by 2.951\% and 2.593\%, respectively, while reducing QRR by 2.350\%. WeFair with only cache retrieval also achieves improvement against the control method. Additional lightweight vector retrieval further reduces the query reformulation rate. All improvements were statistically significant (P<0.01). 
Higher engagement and fewer reformulations validate the practical value of WeFair for meeting users' immediate information needs. \\
\textbf{Online cases analysis.}
Table~\ref{tab:gsb-cases} presents four cases from our online search system. The first illustrates context-guided spelling correction and the next two illustrate intent-aware expansion: adding explanatory and recruitment details brings retrieval results closer to the users' current needs. In the degraded case, adding ``Education Bureau'' provides relevant information, but the rewrite omits ``XX City'', which is needed to identify the intended account. The rewrite query therefore remains incomplete despite being topically relevant. Together with the online gains, these examples support WeFair's effectiveness in translating contextual clues into rewrites that better serve personalized information needs.

\vspace{-5pt}

\section{Conclusion}
We presented WeFair, a fidelity-aware alignment framework for training compact personalized query rewriters that mitigate intent drift. The framework matches human feedback to three complementary roles: demonstration through curriculum SFT, preference learning through DPO, and verification through GRPO. Offline experiments showed that WeFair achieves the best overall accuracy and reduced intent drift while maintaining low inference latency. A 10-day online A/B test further showed significant gains in engagement and fewer query reformulations. Furthermore, our analyses revealed that broad, lower-fidelity context-to-query supervision benefited the complete SFT curriculum, while consumption-based preference learning complemented subsequent verification. In GRPO, strict agreement with human targets yielded higher overall accuracy than the tested auxiliary rewards. These findings support matching supervision fidelity to the role of each training stage. Our method has been deployed in WeChat short-video search, demonstrating the practical value of proposed fidelity-aware alignment for effective personalization under production latency constraints.


\bibliographystyle{ACM-Reference-Format}
\bibliography{reference}
\appendix

\clearpage

\section{Further Implementation Details} \label{sec:imp_details}

\subsection{Adaptation of AWHCP}
\label{app:awhcp-adaptation}
Our local adaptation retains AWHCP's Problem--Intention--Fix--Rewrite (PIFR) structure and SFT--DPO--GRPO sequence~\citep{ou2026awhcp}, while replacing its e-commerce data and feedback services with resources available for WeChat short-video search. Forward examples attach PIFR explanations to existing human reformulation targets, and backward examples are synthesized from real content titles. SFT mixes fast and slow responses without revealing the mode in the input. For DPO, each actor generates fast and slow candidates after a shared problem description. The 2B experiment starts from Qwen3.5-2B, whereas the 8B experiment starts from Qwen3-8B.

The main adaptations are feedback and optimization. A frozen BM25 index of training-visible titles replaces the original product-search service. Its relevance proxy combines lexical similarity to the query and preceding context, without access to the target reformulation. DPO preferences favor larger gains in newly retrieved relevant titles under this proxy. A local Qwen3.8-27B teacher supplies weak labels for the quality judge, replacing the original combination of LLM labeling and expert review. Our output protocol explicitly distinguishes rewriting from preservation, while retaining the original reward weights of 0.3, 0.1, and 0.6 for quality, format, and retrieval increment. The retained DPO configuration uses a learning rate of $10^{-7}$, $\beta=0.5$, and one epoch. GRPO uses batches of 128 queries with eight responses per query, a learning rate of $10^{-6}$, and a KL coefficient of $0.001$. It allows 7680 input and 512 response tokens to accommodate user histories. These archived implementations therefore represent a platform adaptation, with differences in supervision, retrieval feedback, and mode allocation.

\subsection{Training Settings}
\label{app:training-settings}
During SFT training, we use AdamW with a learning rate of $5\times10^{-6}$, weight decay of $0.01$, and an effective batch size of 32 examples. 
For the DPO training, we continue from the SFT policy using preference pairs constructed from user reformulations and subsequent consumption feedback. Each pair shares the same original query and query-time context. For rewrite-required examples, human reformulations or consumption-supported candidates serve as preferred responses. The candidates whose observed retrieval results contain none of the relevant consumed items serve as negative responses. Preservation examples favor the preservation marker over generated rewrites. Missing feedback is excluded, and the subsequent consumption-based behavior feedback is used only to construct supervision. DPO thus learns relative preferences before final GRPO refinement returns to strict human-target verification. Because preference construction relies on sensitive search and consumption histories, we describe its principles while withholding raw user-level records and privacy-sensitive details of the internal feedback-processing pipeline. DPO uses a learning rate of $10^{-6}$, $\beta=0.5$, and an effective batch size of 32 preference pairs.
In GRPO training, this configuration uses a DPO-trained model as its GRPO initial checkpoint. Table~\ref{tab:strict-implementation} provides the hyperparameter settings for the GRPO strict-verification training in WeFair. Advantages are centered within each response group without division by its standard deviation. The KL penalty enters the policy loss separately and does not alter the scalar verification reward. 

\begin{table}[t]
\vspace{5pt}
\centering
\caption{Training hyperparameter settings for our on-policy GRPO training.}
\label{tab:strict-implementation}
\setlength{\tabcolsep}{3pt}
\renewcommand{\arraystretch}{1.06}
\begin{tabular}{@{}m{0.46\columnwidth}m{0.49\columnwidth}@{}}
\toprule
Setting & Value \\
\midrule
Optimizer & AdamW \\
Learning rate / weight decay & $5\times10^{-7}$ / $0.01$ \\
AdamW coefficients & $(0.9,0.999)$ \\
Schedule / warm-up & Constant / none \\
Query batch / PPO mini-batch & 256 / 32 \\
Responses per query & 32 \\
Temperature / top-$p$ & 1.2 / 1.0; no top-$k$ cap \\
Input / response limits & 4,096 / 64 tokens \\
PPO epochs / clipping & 1 / 0.2 \\
KL loss coefficient & 0.02; low-variance estimator \\
Entropy coefficient & 0 \\
Exact / preservation reward & $w_{\mathrm r}=2$, $w_{\mathrm p}=0.5$ \\
Other verification outcomes & 0 \\
\bottomrule
\end{tabular}
\vspace{10pt}
\end{table}

\begin{table}[ht]
\centering
\caption{Training data statistics. K denotes one thousand examples.}
\label{tab:data_stat}
\begin{tabular*}{0.9\columnwidth}{@{\extracolsep{\fill}}lc@{}}
\toprule
Training Data Type & Number \\
\midrule
Context-to-query $\mathcal (C$) & 1500K \\
Human reformulation $\mathcal (R$) & 138K \\
Preservation ($\mathcal P$) & 461K \\
\bottomrule
\end{tabular*}
\vspace{10pt}
\end{table}

\begin{table*}[t]
\centering
\vspace{4pt}
\caption{Environment and decoding settings for the local H20 timing runs. Response limits include reasoning, answer-format markers, and the final answer. The setting can support the latest Qwen-3.8-27B model inference.}
\vspace{2pt}
\label{tab:inference-configuration}
\setlength{\tabcolsep}{4pt}
\renewcommand{\arraystretch}{1.06}
\begin{tabular}{@{}m{0.29\textwidth}m{0.68\textwidth}@{}}
\toprule
Component & Configuration \\
\midrule
\rowcolor{TableSection}[0pt][\tabcolsep]
\multicolumn{2}{@{}l}{\textit{Hardware and software}} \\
GPU / reported memory & NVIDIA H20 / 97,871 MiB per GPU \\
NVIDIA driver & 535.161.08 \\
Python & 3.12.3 \\
PyTorch / CUDA build & 2.13.0+cu130 / 13.0 \\
Transformers / generation backend & 5.10.4 / Hugging Face \texttt{generate()} \\
Weight precision / attention & BF16 / PyTorch scaled dot-product attention (SDPA) \\
\midrule
\rowcolor{TableSection}[0pt][\tabcolsep]
\multicolumn{2}{@{}l}{\textit{Execution and decoding}} \\
Execution mode & Model evaluation mode with \texttt{torch.inference\_mode()} \\
Batch size / concurrent requests & 1 / 1 per GPU \\
Tensor parallel degree / KV cache & 1 / enabled \\
Decoding & Greedy; \texttt{do\_sample=False}, \texttt{num\_beams=1} \\
PyTorch CPU threads / random seed & 1 per process / 42 \\
Warm-up & Three generations per replica and configuration, up to 16 new tokens each \\
Input handling & Full prompts; validate input length plus response budget against the configured context limit \\
Stopping & Tokenizer EOS, \texttt{<|im\_end|>}, or the response limit \\
\midrule
\rowcolor{TableSection}[0pt][\tabcolsep]
\multicolumn{2}{@{}l}{\textit{Prompt-length guard / maximum generated tokens}} \\
WeFair-0.8B/2B/4B, Direct & 4,096 / 64 \\
WeFair-CoT-0.8B/2B/4B & 4,224 / 128 \\
AWHCP-2B & 4,096 / 512 \\
AWHCP-8B & 7,680 / 512 \\
Qwen3.8-27B, Direct and CoT & 8,192 / 8,192 \\
\bottomrule
\end{tabular}
\vspace{10pt}
\end{table*}

\subsection{Details of Data}
\label{app:data-details}
\textbf{Dataset statistics.}
Here, we provide the details and number of training samples used. The training data span four consecutive weeks of search logs. Table~\ref{tab:data_stat} reports the sizes of the three supervision pools. The common test set is collected from the immediately following three days and contains a total of $N=19696$ examples, comprising $N_R=12528$ rewrite-required and $N_P=7168$ preservation-required examples. It follows the training-data filtering pipeline and undergoes additional human review. \\
\textbf{Screening models Setting.}
The Context-to-query screening and the final verification of selected preservation examples use DeepSeek-V4-Flash. The query-to-query human reformulation screening uses three large-size reasoning models Gemini-3.5-Flash, DeepSeek-V4-Flash, and GLM 5.2, while contextual bridging uses three open-source large-size reasoning models: DeepSeek-V4-Flash, Qwen3.5-397B-A17B, and GLM 5.2. Both screening stages use the same majority-voting strategy, in which three models independently assess the same input, and at least two must accept.

\subsection{Online Deployment Details}
\label{app:deployment-details}

This section details the threshold setting and relevance fusion procedure used to incorporate rewritten queries into the online rank system. In our online deployment, we set the relevance-gating threshold to $\tau_{\mathrm{rank}}=0.06$. For input $x=(c,q)$, we obtain the preservation probability from the first decoding step:
\begin{equation}
p_{\mathrm{pres}}(x)
=\pi_\theta(y_1=\bot\mid x)
=\frac{\exp z_{1,\bot}}{\sum_{v\in\mathcal V}\exp z_{1,v}},
\label{eq:deployment-preservation-prob}
\end{equation}
where $z_{1,v}$ is the logit for token $v$, $\mathcal V$ is the full vocabulary, and $\bot$ corresponds to the dedicated \texttt{<|reject|>} token. 

For a candidate video $d$, let $r(q,d)$ and $r(\hat q,d)$ denote its relevance to the original and rewritten queries. When a rewritten query is available, the fused relevance signal supplied to the rank system is
\begin{equation}
r_{\mathrm{fused}}(x,d)=
\begin{cases}
w_{\mathrm{o}}r(q,d)+w_{\mathrm{r}}r(\hat q,d), & p_{\mathrm{pres}}(x)\le\tau_{\mathrm{rank}},\\
r(q,d), & p_{\mathrm{pres}}(x)>\tau_{\mathrm{rank}},
\end{cases}
\label{eq:deployment-relevance-gate}
\end{equation}
where $w_{\mathrm{o}}$ and $w_{\mathrm{r}}$ denote the weights for original-query and rewrite-based relevance. If $\mathrm{pres}(x)$ is above the threshold, we skip computing $r(\hat q,d)$. If the policy outputs the preservation marker, only $r(q,d)$ is used. This fused relevance signal is incorporated into the WeChat short-video search rank system.
To further limit end-to-end latency, we start model prefill when the user taps the search icon and use lightweight vector retrieval for the additional recall channel.

\begin{table*}[t]
\centering
\vspace{4pt}
\caption{Illustrative case rubric. The query is ``crepe cake.'' Recent query-time searches explicitly request cake-eating videos, and the older history contains recipe tutorials. Same-need subsequent behavior confirms an interest in eating videos. No query-time evidence supports a dairy restriction. Hallucination/Intent Drift columns show binary labels.}
\vspace{2pt}

\label{tab:case-rubric}
\setlength{\tabcolsep}{3pt}
\begin{tabular}{@{}
>{\raggedright\arraybackslash}m{\dimexpr0.19\textwidth-1.14\tabcolsep\relax}
>{\centering\arraybackslash}m{\dimexpr0.11\textwidth-0.66\tabcolsep\relax}
>{\centering\arraybackslash}m{\dimexpr0.09\textwidth-0.54\tabcolsep\relax}
>{\raggedright\arraybackslash}m{\dimexpr0.61\textwidth-3.66\tabcolsep\relax}@{}}
\toprule
Candidate & Hallucination & Intent Drift & Interpretation \\
\midrule
Crepe cake eating videos & No & No & Supported refinement consistent with the current need. An equivalent wording may also be acceptable even if it fails Human EM. \\
How to make crepe cake & No & Yes & Recipe history provides provenance, but the candidate switches the task from watching eating videos to learning a recipe. \\
Dairy-free crepe cake recipe & Yes & No & The unsupported dietary constraint places this candidate in HaR. Although the task also changes, IDR is reserved for context-supported rewrites. \\
Crepe cake & N/A & N/A & When treated as preservation, this output fails the action decision and Judge Recall on a rewrite-required example, despite introducing no unsupported content. \\
\bottomrule
\end{tabular}
\vspace{10pt}
\end{table*}

\begin{table*}[!t]
\centering
\caption{WeFair's inference instruction for evaluation, with an English translation and anonymized context fields. The context layout is simplified for presentation.}
\vspace{2pt}

\label{tab:rewriter-prompts}
\setlength{\tabcolsep}{6pt}
\renewcommand{\arraystretch}{1.12}
\begin{tabular}{@{}
>{\centering\arraybackslash}m{\dimexpr0.47\textwidth-\tabcolsep\relax}
>{\centering\arraybackslash}m{\dimexpr0.53\textwidth-\tabcolsep\relax}@{}}
\toprule
\textbf{Chinese} & \textbf{English translation} \\
\midrule
\rowcolor{TableSection}[0pt][\tabcolsep]
\multicolumn{2}{@{}c}{\textit{\textbf{Evaluation-time inference instruction}}} \\
\casezh{仅输出：上下文明确支持补全、纠错或个性化改写，且不偏离当前搜索意图时，用}\texttt{<|qr|>}\casezh{改写query；否则用}\texttt{<|reject|>}\casezh{。}
& Only output \texttt{<|qr|>} followed by the rewritten query when the context clearly supports query completion, correction, or personalized rewriting without deviating from the current search intent; otherwise, output \texttt{<|reject|>}. \\
\addlinespace
\rowcolor{TableSection}[0pt][\tabcolsep]
\multicolumn{2}{@{}c}{\textit{\textbf{Personalized Input fields}}} \\
\textless\casezh{浏览记录}\textgreater\par
\textless\casezh{相关历史搜索}\textgreater\par
\textless\casezh{近期搜索记录}\textgreater\par
\textless\casezh{搜索点击记录}\textgreater\par
\textless\casezh{其他查询时上下文}\textgreater\par
\textless\casezh{当前查询}\textgreater
& \textless Browsing history\textgreater\par
\textless Related previous searches\textgreater\par
\textless Recent searches\textgreater\par
\textless Search-click history\textgreater\par
\textless Other query-time context\textgreater\par
\textless Current query\textgreater \\
\bottomrule
\end{tabular}
\vspace{10pt}
\end{table*}

\subsection{Fidelity Errors Interpreting }
\label{app:fidelity-interpretation}
In order to better clarify the intent drift problem. We provide a case(in Table~\ref{tab:case-rubric}) to illustrate why contextual support, intent alignment, and a correct rewriting decision must be assessed separately. This example has the original input query ``crepe cake''. Recent user searches and views history support an interest in eating videos, whereas older history contains lunch recipe tutorials. Adding ``eating videos'' resolves the current need. Adding ``how to make'' has contextual support but changes the task, illustrating intent drift despite topical relevance. A dairy-free restriction introduces an additional requirement unsupported by the available context, defining as hallucination rewrite.

Although an output can both add unsupported information and change the search goal, the reported metrics use distinct categories: unsupported additions contribute to hallucination, while intent drift is reserved for context-supported rewrites inconsistent with the current need. Therefore, the dairy-free recipe in the example contributes to hallucination rewrite even though it also redirects the task. Preserving the original query introduces neither error, but still fails the decision when rewriting is required. Low hallucination or intent drift alone therefore does not establish successful intent recovery. Conversely, a faithful alternative expression can pass semantic evaluation while failing exact matching. These distinctions motivate evaluating rewrite success and preservation alongside fidelity errors, and interpreting strict verification as a conservative training signal to align human needs.

\vspace{5pt}

\subsection{Auxiliary Reward Setting}
\label{app:rm-implementation}
In Section~\ref{sec:reward-analysis}, semantic similarity, subsequent consumption feedback, and reward-model judgments. Similarity-Augmented Verification measures semantic similarity using the character-level F1 score between the candidate rewrite and the human reformulation, scaled by $0.5$. For Consumption-Augmented Verification, we follow the retrieval-based criteria used in DPO preference construction. A candidate receives an auxiliary reward of $0.5$ if its top-10 retrieval results contain at least one of up to three subsequent relevant consumed videos, and zero otherwise.

For reward-model feedback, we train a 4B Qwen-based judge to assess whether a candidate preserves the intended refinement, accepting equivalent wording while rejecting intent changes or missing constraints. RM-Augmented Verification assigns an auxiliary reward of $0.5$ when a single judgment accepts the candidate, and zero otherwise. RM-Augmented Verification(4 Vote) instead requires full acceptance across four independently sampled judgments to assign the same reward. These judgments use temperature $0.7$, top-$p$ $0.95$, and a 64-token response limit. Both RM variants restrict judging to well-formed, non-exact candidates that differ from the original query and have a rewrite target.



\subsection{Inference Efficiency Measurement}
\label{app:efficiency-protocol}
We selected the software stack for compatibility with Qwen3.8, enabling evaluation of the query-rewriting capabilities of the latest medium-sized LLMs. For consistent timing comparisons, all local models used the default Transformers generation path without additional inference-framework optimizations. Precision and decoding parameters are explicitly specified below. Table~\ref{tab:inference-configuration} summarizes the hardware, software, and decoding settings of the local H20 timing runs. Eight independent model replicas process disjoint subsets of the timing queries, with each replica occupying one GPU and serving requests sequentially. Notably, the setting is not the same as our online inference system.

\section{Inference Prompt}
\label{app:prompts}
Table~\ref{tab:rewriter-prompts} presents the evaluation-time inference instruction used for direct WeFair, together with anonymized input fields and English translations. The model receives the original query and query-time context, then outputs either \texttt{<|qr|>} followed by a rewrite or \texttt{<|reject|>} to preserve the query, without generating reasoning traces. Both action markers are added to the training vocabulary as dedicated tokens, so each occupies one token position. Placeholders replace user-specific content, and the context layout is simplified for presentation.


\end{document}